# Artificial intelligence contribution to translation industry: Looking back and forward

Mohammed Q. Shormani,
Department of English Studies, Ibb University, Ibb, Yemen
Email: shormani@ibbuniv.edu.ye
Yehia A. Al-Sohbani,
Open Arab University, Riyadh, Saudi Arabia,
Email: y.alsohbani@arabou.edu.sa

**Abstract**
This study provides a comprehensive analysis of artificial intelligence (AI) contribution to research in the translation industry (ACTI), synthesizing it over forty-five years from 1980-2024. 13220 articles were retrieved from three sources, namely WoS, Scopus, and Lens; 9836 were unique records, which were used for the analysis. we provided two types of analysis, viz., scientometric and thematic, focusing on Cluster, Subject categories, Keywords, Bursts, Centrality and Research Centers as for the former. For the latter, we provided a thematic review for 18 articles, selected purposefully from the articles involved, centering on purpose, approach, findings, and contribution to ACTI future directions. This study is significant for its valuable contribution to ACTI knowledge production over 45 years, emphasizing several trending issues and hotspots including *Machine translation, Statistical machine translation, Low-resource language, Large language model, Arabic dialects, Translation quality,* and *Neural machine translation*. The findings reveal that the more AI develops, the more it contributes to translation industry, as Neural Networking Algorithms have been incorporated and Deep Language Learning Models like ChatGPT have been launched. However, much rigorous research is still needed to overcome several problems encountering translation industry, specifically concerning low-resource, multi-dialectical and free word order languages, and cultural and religious registers.

## 1. Introduction

Machine translation has been led by artificial intelligence (AI) revolution, which demarcates its developments over time since its inception. The study problem lies in the paucity of research that addresses the question of whether AI contributes to research in translation industry (ACTI), reviewing its knowledge production for a considerable period of time. As this article unveils, AI contribution to translation industry is considerably rigorous, resulting in machine translation (MT) industry. AI refers to creating, modeling and/or producing computer intelligence like that of human. AI started shaping its foundations in 1955 (Minsky, 1961; Turing, 1950). It is the development of computer models that can perform tasks like human intelligence. These computer models involve algorithms trained on large datasets to learn patterns and make predictions (McShane & Nirenburg, 2021). AI aims to simulate intelligent behavior including learning, problem-solving, perception, and even decision-making (Liao et al., 2018). These models are reported to perform tasks with considerable accuracy (Linzen & Baroni, 2021; Gulordava et al., 2018). The great revolution led by AI generative models depends on the use of the Neural Networking Algorithms (NNAs) (Peng et al., 2023). These algorithms have been utilized in AI models. For example, "the power of GPT to provide interactive and dynamic responses, mimicking human-like conversation" has been utilized in these models (Sohail et al. 2023, p. 1).

Scientometrics is considered one of the modern tools to provide invaluable insights into how research in a field of study develops, demarcating the strengths and weaknesses, knowledge gaps and future directions. It is defined as "the study of the quantitative aspects of science and technology seen as a process of communication", used across a wide range of research areas including scientific and social sciences and humanities disciplines (Mingers & Leydesdorff, 2015, p. 1). It has been used extensively in characterizing such trends utilizing statistics, mathematics, software-generating visualization devices to conceptualize and concretize the achievement of scientific research. It covers a wide range of disciplines to find out to what extent a particular field of knowledge has achieved distribution and how the scientific community has engaged in this field (Geng et al., 2024). Recently, two influential software have been developed, namely CiteSpace and VOSviewer, which are used to conduct such types of scientometric studies, covering a wide range of disciplines including scientific and social sciences (see also Sooryamoorthy, 2020).

In this study, we provide a comprehensive analysis of the AI contribution to research in the translation industry in a period spanning more than four decades, investigating its beginnings, developments over time, and unveiling the current intellectual landscape, (re)emergent trends and hotspots. To the best of our knowledge, no previous study has tackled this phenomenon, specifically focusing on scientometric and thematic analyses, hence this study serves to fill this gap. Our analysis employs Cluster Analysis, Bursts, (betweenness) Centrality, Subject Categories, Document Co-citation Analysis (DCA), Research Centers and Keywords. These metric indicators unveil ACTI intellectual landscape, trending issues and hotspots, demarcating the strengths and weaknesses of ACTI, and pinpointing the possible knowledge gaps. These scientometric analyses were followed by a thematic analysis of 18 articles carefully chosen based on two criteria, namely i) the most citing articles and experimental articles relating to AI, Computational Linguistics (CL) and Natural Language Processing (NLP), that provide valuable insights into how MT could be developed to overcome the existing problems. And ii) these featured an in-depth, qualitative exploration of key themes emerging from representative contributions in ACTI

Thus, the rest of article is organized as follows. In section 2, we articulate the literature review, discussing studies on ACTI. In section 3, we spell out the study methods, distilling our data sources, data screening, and methods of analysis. In section 4, we present the results and in section 5 we discuss these results. In section 6, we conclude the article, outlining some actionable practices and future frontiers of ACTI research.

## 2. Literature review

### 2.1. Artificial intelligence

The contribution of artificial intelligence to translation industry goes beyond one research article, as AI accelerates translation, contributing to its success, though problems arise now and then concerning some typological differences between languages as we will see in this article. Through AI, translation has entered a technological world, witnessing a huge shift from just a human endeavor to a "computer" craft, due mainly to the developments technology, internet and AI have undergone. In the current time, translation industry depends on AI translation tools and apps, and the human role becomes limited to just post-editing, specifically regarding noncultural texts (Shormani, 2024b). Put differently, AI involvement in translation industry results in several web apps including Google Translate and Microsoft Translate, which are basically Statistical Machine Translation (SMT) devices (see e.g. Farzi & Faili, 2015). With the advancements of AI technology, NNAs dramatically change AI orientations leading to launching several Language Learning Models (LLMs), the most powerful is perhaps ChatGPT. As for translation, ChatGPT can translate massive amounts of data from any language into another in a very short time, performing competitively better than any nonneural translation tool, specifically ChatGPT-4 (Jiao et al., 2023), due to the huge amounts of internet data ChatGPT has been trained on (Kung et al., 2023; Larroyed, 2023; Shormani, 2024b &c).

Recently, OpenAI has developed language models based on NNAs including ChatGPT which "can generate content across various domains, such as text, images, music, and more" (Ray, 2023, p. 121). It has not only impacted these aspects, but also leveraged the "scientific research, spanning from data processing and hypothesis generation to collaboration and public outreach" (Ray, 2023, p. 121). ChatGPT is an example of large language models, which creates much fears among human translators due its good quality of translation both in content and meaning being more accurate than other nonneural tools such as Google Translate (Kenny, 2022; Kung et al., 2023; Lee, 2023; Ray, 2023; Siu, 2023; van Dis et al., 2023).

### 2.2. Machine Translation

Translation is a process and product; it could be viewed as a transmitter of our knowledge, experiences, and ideologies from a nation to another, from ancestors to descendants (James, 2002; Shormani, 2020, 2024b; Shormani & Alfahd, 2025). Translation is not limited to just transferring the meaning of a text wording, but it involves creativity and innovation (Dugonik et al., 2019). MT has been one of the major concerns of AI, CL and NLP specialists, involving linguistic texts that can be produced and read by computer, and employing "methods for extracting linguistically valuable information from such texts" (Brown et al., 1993, p. 263). It started with employing rule-based mechanism, then statistics, and finally NNAs, coming up with several types of translation mechanism (de Coster et al., 2023), and reflecting the stages it has passed through in each of these types. There are several spheres of MT including TV and film industry like subtitling, dubbing, subtitling, and lexigraphy and dictionaries (Fernández-Costales et al., 2023; Zhang & Torres-Hostench, 2022).

It is estimated that MT history spans four centuries now, referring to René Descartes in 1629 who held that what can be expressed in a language can be expressed in another language sharing one symbol (Yang et al., 2020). However, the actual beginning of MT as an industry was in 1960s, recommended by several scholars in the first International Conference on Machine Translation in 1952 (Forcada et al., 2011). SMT started as translating single/isolated words, but since then it has developed to include phrases and sentences. MT has witnessed tremendous developments since then shaping its frontiers into three types: Rule-based Machine Translation (RMT), Statistical Machine Translation, and Neural Machine Translation (NMT) (Koehn et al., 2003; Forcada et al., 2011).

The main idea of RMT is that a word in L1 can have an equivalent in L2. In this sense, what MT does is replace L1 word with L2 word in a syntactic-based manner. However, languages differ in the way syntax works in each language. Put simply, languages such as English follow an SVO word order, but languages like Arabic follow a VSO word order, while languages like Hindi are SOV languages. Additionally, English, for instance, is a head -first language, while languages like Hindi are head-last languages, as in *in the house* and *ghar mae*, respectively. These syntactic and typological variations were not easy to master by machine (precisely computers), and hence MT output was not always satisfactory. Another problem encountering RMT was semantic in nature. For example, polysemy, which means that a word can have several meanings in L1, that may not be possible to be captured by an RMT device.

RMT's drawbacks were the starting point to think of an alternative, which is SMT at that time. SMT's main idea is that a word can have multiple meanings and computers just identify the best match from a bilingual corpus based on statistics. The bilingual corpus could be thought of as a bilingual lexicon, *WordNet* is the best example. This idea is behind Google Translate working mechanism, where machines identify the best match/meaning of a word in L1 from L2 as they are iterated in the bilingual corpus. SMT moves translation far strides, underscoring the translation industry, and researching it spanned a long time, as will be clear in the review of articles in Table 8. However, SMT's produced translations were also not satisfactory as can be seen in Google Translate's output. And with the huge developments of technology and AI, MT enters a new world, its best manifestation is incorporating neural algorithms, or deep language learning networks. This results in what is known as neural machine translation, which will be the focus of the section to follow.

NMT is mainly based on NNAs, ascribed to the huge developments taking place in AI technology (Peng et al., 2023). These ideas started to crystallize in the early 1990s, where "intelligence" invaded computer industry (Pollack, 1990). However, the starting point of NNAs emerged in about 2010 with the advent of Deep Learning Language Models (DLLMs). Several DLLMs have been developed including hybrid models based on SMT and NMT (Dugonik et al., 2019), multi-NMT models (Huang et al., 2023), a transformer model (Baniata et al., 2021), human-post-editing model (Formiga et al. (2015), and AI-human models (Li et al., 2023). With the help of NLP and AI, NMT has been a surge in translation industry, attracting ample scholars and researchers from linguistics, computer science, technology, computer engineering, programing, CL, and NLP. The most recent development in this regard is ChatGPT, the purpose of launching which was translation, but eventually it has been used for many and various purposes, as we have seen so far.

There are some studies that have tackled reviewing AI and translation research, though different from our study in purpose, scope and content, specifically the methodologies employed, number of articles reviewed, and themes focused on. For example, Mohamed et al. (2024) analyze the evolving role of AI in translation, tracing it from RMT to NMT, and highlighting how these have significantly improved translation quality. However, the paper also identifies persistent challenges, and research gaps, particularly in incorporating "cross-lingual dialect adaptability and the advancement of Artificial Intelligence translation systems, with a focus on prioritizing inclusion and cultural understanding." (p. 25553). However, the number of articles reviewed in this study is not explicitly stated. Additionally, Genovese et al. (2024) evaluate the effectiveness of AI in clinical language translation and interpretation. The study analyzes nine peer-reviewed articles from 2019 to 2024, focusing on AI tools such as Google Translate, Microsoft Translator, Apple iTranslate, and the Asynchronous Telepsychiatry (ATP) App. Their findings indicate that AI translation tools demonstrate high accuracy (83–97.8%) when translating from English, but lower accuracy (36–76%) when translating into English. Clinicians expressed concerns regarding the reliability and quality of AI translations, often using these tools as a last resort or for brief communications. The review concludes that while AI shows promise in clinical translation, specifically for short interactions, the complexity of medical consultations necessitates a balanced approach that combines AI tools with human translation services to ensure register-specific quality. Thus, this study strives to answer the following questions:

1. What are the most salient AI contributions to research in translation industry in 1980-2024 and 2015-2024 timeframes?
2. How do scientometric indicators represent ACTI knowledge production in each timeframe, and what are the differences between them?
3. What are the main themes of ACTI knowledge production, and what are the future frontiers of this area of study?

## 3. Methodology

In this section, we present the methods adopted in this study concerning the data collection and search strategies and terms used. We also outline the methods of the data screening process, data export and conversion, and methods of analysis.

### 3.1. Data collection: search strategies and terms

Our purpose in this article is to provide a comprehensive account of ACTI intellectual production between 1980 and 2024. Hence, we aim to collect as many articles as possible from trustworthy sources. Three reliable databases, namely Web of Science(WoS) Scopus and Lens, were chosen, as being the most sourced databases used by scholars in several and varied scientometric studies (see e.g. Chen, 2006; Sooryamoorthy, 2020). The data collection was performed on March 27, 2024, from the three databases at the same day to avoid data overlap. The first thing in data collection was setting the timespan in the three search engines to 1980-2024. In WoS, no publication was found before this 1980. These sources' search engines were also set to exclude nonarticles. Nonarticles will be stated in **section 3.2** below. Table 1 summarizes search query and terms, results and the data retrieved.

**Table 1: Query, search types, search terms across Lens, Scopus and WoS**

| Query | Search type | Research terms (Searched on April 2, 2024) | Lens | Scopus | WoS | Total |
|---|---|---|---|---|---|---|
| nonspecific | Title-AB-KW | Artificial intelligence" OR "Computer" OR "Machine Translation" OR "Neural Algorithms" OR "Neural Networks" OR Computational Linguistics" OR "ChatGPT" AND "Translation" OR "Post-editing" OR "Human Translation" | 6016 | 1219 | 5332 | 12567 |

| | Limits. & excld. Data | (LIMIT-TO ( LANGUAGE, “English” ) ) AND ( EXCLUDE ( NONRESEARCH-ARTICLES) ) ) | | | | |
|---|---|---|---|---|---|---|
| specific | Title | Artificial intelligence” OR “Computer” OR “Machine Translation” OR “Neural Algorithms” OR “Neural Networks” OR Computational Linguistics” OR “ChatGPT” AND “Translation” OR “Post-editing” OR “Human Translation” | 338 | 173 | 142 | 653 |
| | Limits. & excld. Data | (LIMIT-TO ( LANGUAGE, “English” ) ) AND ( EXCLUDE ( NONARTICLES) ) ) | | | | |
| Total | | | 6354 | 1392 | 5474 | 13220 |

### 3.2. Data screening

The search strategies result in 13220 in total. CiteSpace and Mendeley were used to remove duplicates, i.e. articles that occur in more than one database. The removal process results in 9836 unique records and 3384 duplicate articles. The search engines of WoS, Scopus, and Lens were utilized to exclude nonarticles while searching. Nonarticles include meeting abstracts, enriched cited references, book reviews, review articles, editorial materials, open publisher-invited review, retracted publications, corrections, letter, and notes.

### 3.3. Data export and conversion

The nature of our data is multi-sourced, viz., WoS, Scopus, and Lens. These three sources each have different file format to export the data. So, our data were exported in various formats as displayed in Table 2.

**Table 2: Data source, file format, analysis and software used**

| Data source | Export Format | Analysis | Software used |
|---|---|---|---|
| Lens | CSV | Bibliometric Analysis | VOSviewer/ CiteSpace |
| Scopus | CSV | Bibliometric Analysis | VOSviewer/ CiteSpace |
| WoS | Endnote Desktop | Bibliometric & Scientometric Analyses | CiteSpace/ VOSviewer |

After exporting our data, CiteSpace algorithms were used to convert Lens and Scopus data to WoS-to-be-processed data. After the conversion process, the data were compiled in one folder for CiteSpace output analysis. As mentioned above, the net data comprise 9836 articles (from the three sources).

### 3.4. Methods of analysis

Two methods were employed to analyze our data, scientometric analysis and thematic analysis of ACTI research in forty-five years. The former concerns analyzing Clusters/trends, (betweenness) Centrality, Bursts, and Silhouette. Both CiteSpace and VOSviewer were used for these types of analysis. We adopted these concepts from Chen (2006) and Chen et al. (2019) to conduct the scientometric analysis. Scientometric analysis is a modern tool; it provides substantial insights into how ACTI develops, demarcating its strengths and weaknesses, knowledge gaps and future directions. Thematic analysis involves analyzing some articles, carefully selected from the articles involved in our study. They were chosen considering the Cluster analysis, precisely from the major citing articles of each cluster, in addition to considering other analyses such as Cluster/trend, Centrality, Subject categories, Keywords, and Bursts, coming up with 18 articles. We adopted some notions and techniques from Khosravi et al. (2024) to conduct the thematic analysis focusing on such areas as author, year of publication, purpose/aim, methodology, findings, and most importantly the articles' contribution to ACTI future directions. These areas allow us to come up with several themes (Table 8).

Given our purposes and questions, [CiteSpace](#) 6.3.1R, and [VOSviewer](#), two scientometric programs, were utilized to analyze and generate mapping visualizations of ACTI research. We also utilize the bibliometric indicators available in WoS, Scopus, and Lens including authorship, affiliation, article title, keywords, cited and citing references, cited reference count, publication year, country, and language. Quantitative and qualitative approaches were adopted adopting analytic and descriptive methods. The quantitative approach concerns the data collected in terms of scientometric analyses including Cluster/trend, Author Keywords, Subject Categories, Co-citation or DCA analyses, etc. The qualitative approach, however, concerns the thematic analysis of 18 research articles. It is also embodied in the discussion section, where we interpret the quantitative findings placing much emphasis on the thematic review and on how to implement the study results on encountering the problems arising specifically from low-resource and multi-dialectical languages.

**4.3.1. ACTI development over time**

Fig 1 displays ACTI knowledge production along with g-index over the 1980-2024 timeframe (extracted from CiteSpace 6.3.1R). It shows that ACTI publication was not stable, specifically between 1980 and 2004, i.e. during this period ACTI publications sometimes decrease and some other times increase. For instance, in 1980, there were 12 publications, then the number of publications increased, but again decreased to 9 in 1986. It reached 132 in 1999, but again decreased to only 13 publications in 2001. However, from 2005 onwards it increases considerably. However, this increase is gradual over time. For example, in 2005 the number of publications is 121, reaching the highest number in 2023 with 1309 publications. Note that there are only 240 publications in 2024, but this does not mean that ACTI research decreases, but rather this number of publications is only in 3 months (January-March).

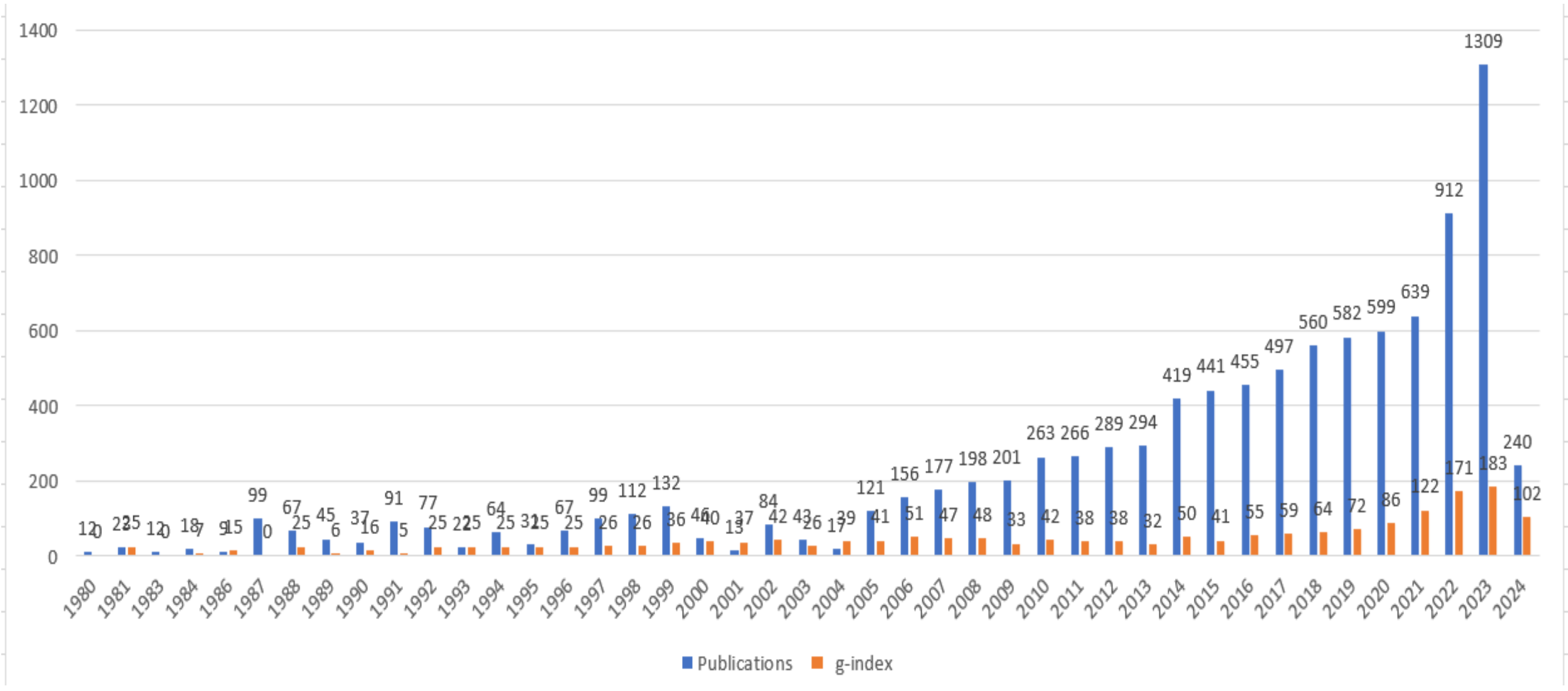

*Fig 1: ACTI publications and g-index over time from 1980 to 2024*

## 4. Results

### 4.1. Cluster analysis

In this section, we present cluster analysis in our study, pinpointing ACTI research in two timeframes, viz., 1980-2024 and 2015-2024 timeframes. We aim to see subtle shifts and the impact of technological breakthroughs on MT industry.

#### 4.1.1. 1980-2024 timeframe

Cluster analysis represents ACTI knowledge production, trends and hotspots. In this timeframe, there are 471 research clusters/trends as generated by CiteSpace, but we modified CiteSpace settings to visualize only 12 top clusters.

**Table 4: Cluster analysis in terms of Size, Silhouette LLR and average year (from old to new)**

| ClusterID | Label (LLR) | Size | Silhouette | Average Year |
|---|---|---|---|---|
| 4 | small parallel corpora | 76 | 0.988 | 2004 |
| 13 | wide-coverage multilingual semantic network | 19 | 0.987 | 2009 |
| 3 | statistical machine translation | 95 | 0.974 | 2011 |
| 8 | machine translation | 39 | 0.971 | 2011 |
| 2 | neural machine translation | 100 | 0.914 | 2016 |
| 6 | deep learning | 48 | 0.941 | 2016 |
| 0 | generative adversarial network | 125 | 0.96 | 2017 |
| 1 | low-resource language | 117 | 0.883 | 2018 |
| 5 | large language model | 49 | 0.945 | 2020 |
| 15 | arabic dialects | 16 | 0.988 | 2020 |
| 7 | Multi-modal neural machine translation | 40 | 0.979 | 2021 |
| 14 | deep learning models | 16 | 1 | 2023 |

Table 4 showcases Cluster analysis of ACTI research in 1980-2024 timespan, providing profound insights into how ACTI research develops over forty-five years, uncovering ACTI intellectual landscape and trending issues. These clusters are sorted by the emerging date (i.e. the average year of emergence). Intellectual landscape of ACTI in forty-five years is represented by the top 12 clusters as displayed in Table 4. Cluster 4 *Small parallel corpora,* emerging around 2004, marks the scholarly community's interest in ACTI in that time. Having 76 articles and a Silhouette value of 0.998 is prominent, underscoring ACTI specialists' interest in using computer technologies to compose Corpora like *WordNet*. Around 2009, Cluster 13 *Wide-coverage multilingual semantic network* came to play, constituting a trending issue in ACTI, though having 19 articles. Cluster 3 S*tatistical machine translation* evolves around 2011, preceding *Machine translation*, which evolved around 2011, in the same year. Although MT began to take shape in 1960, machine translation in our corpus came to play around 2011, thus mirroring the research focus of ACTI scholars. The three last Clusters, namely *Arabic dialects*, *Deep language model,* and *Deep learning models* evolve around 2020, 2021 and 2023, respectively, demarcating ACTI's focus in these periods. Cluster 13 *Wide-coverage multilingual semantic network* contains the least number of articles, i.e. 19 articles. It emerged around 2009 and represents a niche yet notable trend in ACTI research during that period. However, Cluster 0 *Generative Adversarial Network,* emerging 2017 and comprising 125 articles, is identified as the top prominent trending issue in ACTI research over the forty-five-year timespan.

### 4.1.2. 2015-2024 timeframe

In this section, we present Cluster analysis of ACTI trends and hotspots (re)emerging between 2015 and 2024, i.e. in a 10-year period. This timeframe contains 214 clusters. Our aim here is to examine to what extent ACTI knowledge production in this timeframe differs from that produced in 1980-2024 timeframe (Table 4). Table 5 summarizes Cluster analysis of the top 12 clusters in 2015-2024 timeframe.

**Table 5: Cluster analysis in terms of Size, Silhouette LLR and average year (from old to new)**

| ClusterID | Label (LLR) | Size | Silhouette | Average Year |
|---|---|---|---|---|
| 3 | statistical machine translation | 75 | 0.996 | 2015 |
| 18 | sequence-level learning | 23 | 0.986 | 2015 |
| 4 | deep learning | 69 | 0.921 | 2016 |
| 9 | convolution-enhanced evolving attention network | 27 | 0.995 | 2017 |
| 2 | neural machine translation | 99 | 0.903 | 2018 |
| 5 | document-level neural machine translation | 48 | 0.937 | 2018 |
| 1 | low-resource language | 125 | 0.945 | 2018 |
| 7 | machine translation | 37 | 0.944 | 2019 |
| 0 | generative adversarial network | 138 | 0.864 | 2019 |
| 10 | deep neural machine translation | 20 | 0.947 | 2020 |
| 6 | multi-modal neural machine translation | 38 | 0.982 | 2021 |
| 11 | function graph call context | 19 | 0.988 | 2023 |

Table 5 presents the Cluster analysis of ACTI research over a 10-year period from 2015 to 2024. The clusters are arranged based on their average year of emergence. Within this timeframe, the first three clusters are *Statistical machine translation*, *Sequence-level learning* and *Deep learning* emerging in 2015 and 2016, Clusters 3, 18, and 4, respectively. The three last ACTI trends emerged in the 2020-2023 period. These are *Deep neural machine translation, Multi-modal neural machine translation*, and *Function graph call context*, Clusters 10, 6, and 11, respectively. Cluster 11, includes the least number of articles (19 articles). It emerged in 2023, having the highest Silhouette value, i.e. 0.988, indicating strong internal consistency and thematic coherence. However, Cluster 0 *Generative Adversarial Network* (GAN), containing 138 articles. As the most prominent trending topic in ACTI research over 6 years, this cluster underscores the significant influence of GANs on AI and AI-driven language processing, with a respectable Silhouette value of 0.864, highlighting both its popularity and internal thematic strength. Fig 2 displays the Cluster network and Fig 3 the Timeline view of this timeframe.

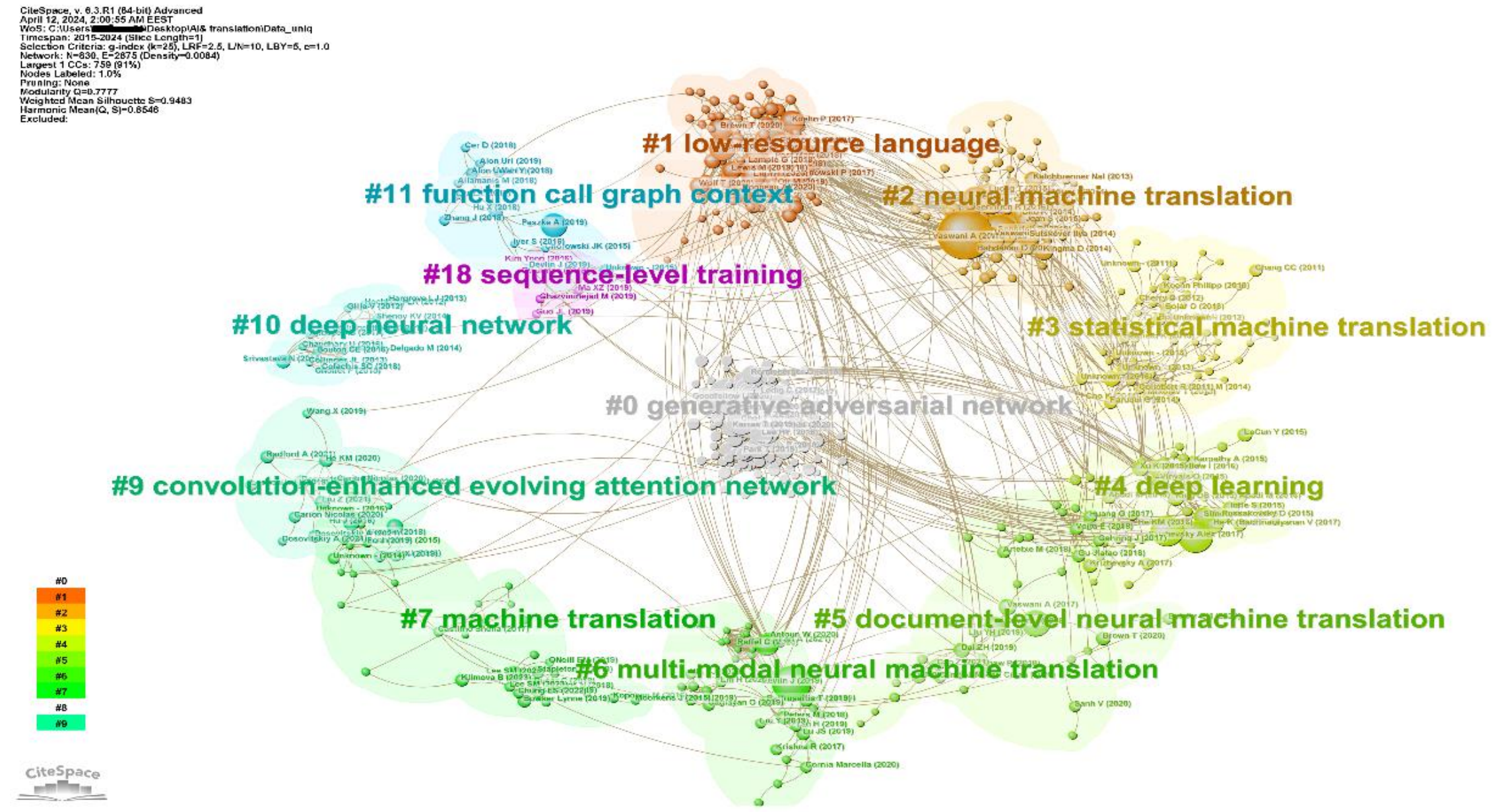

*Fig 2: Cluster network of 2015-2024 timeframe*

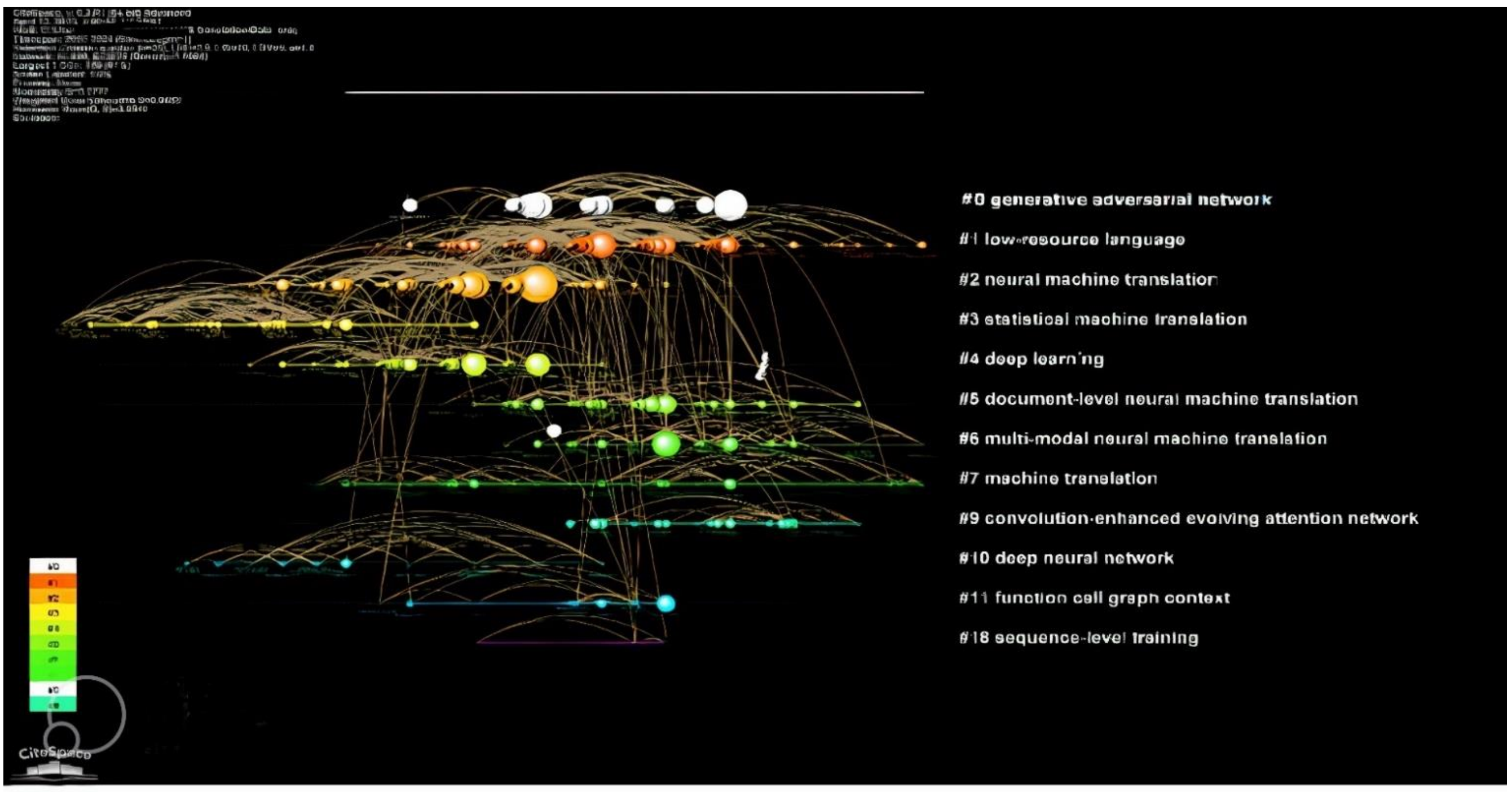

*Fig 3: Timeline view of 2015-2024 timeframe*

### 4.2. Bursts

The term “Bursts” in scientometric analysis refers to a frequency surge of a citation of an article in a particular period of time (Chen, 2006; Ballouket al., 2024). In what follows, we devote a space to discussing subject categories and author’s keywords based on Bursts. It should be noted that all these analyses are based on 1980-2024 timeframe.

#### 4.2.1. Subject categories

During forty-five years, ACTI research has witnessed tremendous developments starting from RMT, SMT, NMT, utilizing several web apps such as Google Translate, Microsoft Translate, and finally DLLMs such as ChatGPT. It has also interacted with and been contributed to by several areas including language/linguistics, computer science, physics, and artificial intelligence, resulting in many and various research areas, application models such as devising semantic networks, small corpora, software engineering, and ChatGPT. All these moves in ACTI research lead to several attempts to seek answers to several questions including: i) is it possible to apply these developments across languages like Arabic, Spanish, Chinese, ii) what is the reality of the resultant translations by applying ACTI technologies and discoveries, and iii) do MT's translation products need human role, and if so, to what extent? These questions, among others, are currently leading ACTI worldwide. However, each question arises in a different time as showcased by "year" in Fig 4, demarcating scholarly community ACTI orientations and interests in finding out answers to each question in (i-iii), as indicated by the beginning of the Bursts period, and mapping out when scholarly community's interest/research focus moves from one question to another, as revealed by the ending period of Bursts.

**Top 10 Subject Categories with the Strongest Citation Bursts**

| Subject Categories | Year | Strength | Begin | End | 1980-2024 |
|---|---|---|---|---|---|
| statistical machine translation | 1999 | **40.5** | 2003 | 2019 | |
| linguistics, semantic networks | 2003 | **20.28** | 2006 | 2017 | |
| language, small corpora | 1992 | **19.92** | 2006 | 2017 | |
| computer science, software engineering | 1984 | **17.24** | 2007 | 2015 | |
| neural machine translation, arabic dialect | 2000 | **15.66** | 2019 | 2022 | |
| translation quality | 2011 | **14.25** | 2012 | 2019 | |
| physics, applied | 2011 | **12.94** | 2021 | 2022 | |
| new approach, post-editing | 2010 | **12.41** | 1995 | 2009 | |
| artificial intelligence, deep learning | 2016 | **11.7** | 2022 | 2024 | |
| deep learning model, chatgpt | 2020 | **11.65** | 2022 | 2024 | |

***Fig 4: Top 10 subject categories with the Strongest Citation Bursts***

### 4.2.2. Keywords

Fig 5 presents the top used author's keywords based on the strongest citation Bursts in ACTI production for forty-five years. *Statistical machine translation* unveils strong focus on ACTI production. It emerges around 1999, ranking first with 55.24 strength beginning in 2003 and ending in 2019, and lasting about 16 years. *Information retrieval* emerges in 1997 with strongest Bursts of 17.96 lasting for 22 years, from 1997 to 2019. Then the focus moves to *Support vector machine* which emerges as a keyword in 2006, its strength lasted about 13 years from 2006 to 2018. *Translation quality* maps out the important focus on how translation should be, demarcating this interest for 8 years, 2011-2019. *Target language*, as a key focus word in ACTI, with Bursts strength of 15.15 lasted for 19 years, 1998-2016, underscoring the importance of target language in the translation process. The top 10 keywords end with *Cross-language information retrieval* with Bursts strength of 12.37, emerging in 1998, and spanning from 1998 to 2014.

**Top 10 Keywords with the Strongest Citation Bursts**

| Keywords | Year | Strength | Begin | End | 1980 - 2024 |
|---|---|---|---|---|---|
| statistical machine translation | 1999 | **55.24** | 2003 | 2019 | |
| information retrieval | 1997 | **17.96** | 1997 | 2019 | |
| support vector machine | 2006 | **16.65** | 2006 | 2018 | |
| translation quality | 2004 | **15.53** | 2011 | 2019 | |
| target language | 1998 | **15.15** | 1998 | 2016 | |
| new approach | 1995 | **15.06** | 1995 | 2009 | |
| support vector machines | 2003 | **14.99** | 2003 | 2018 | |
| machine translation | 1984 | **13.76** | 1998 | 2006 | |
| machine translation system | 1998 | **12.99** | 1998 | 2006 | |
| cross-language information retrieval | 1998 | **12.37** | 1998 | 2014 | |

***Fig 5: Top 10 keywords with the Strongest Citation Bursts***

For a full picture of keywords, consider Fig 6 displaying keywords' Density view generated by VOSviewer.

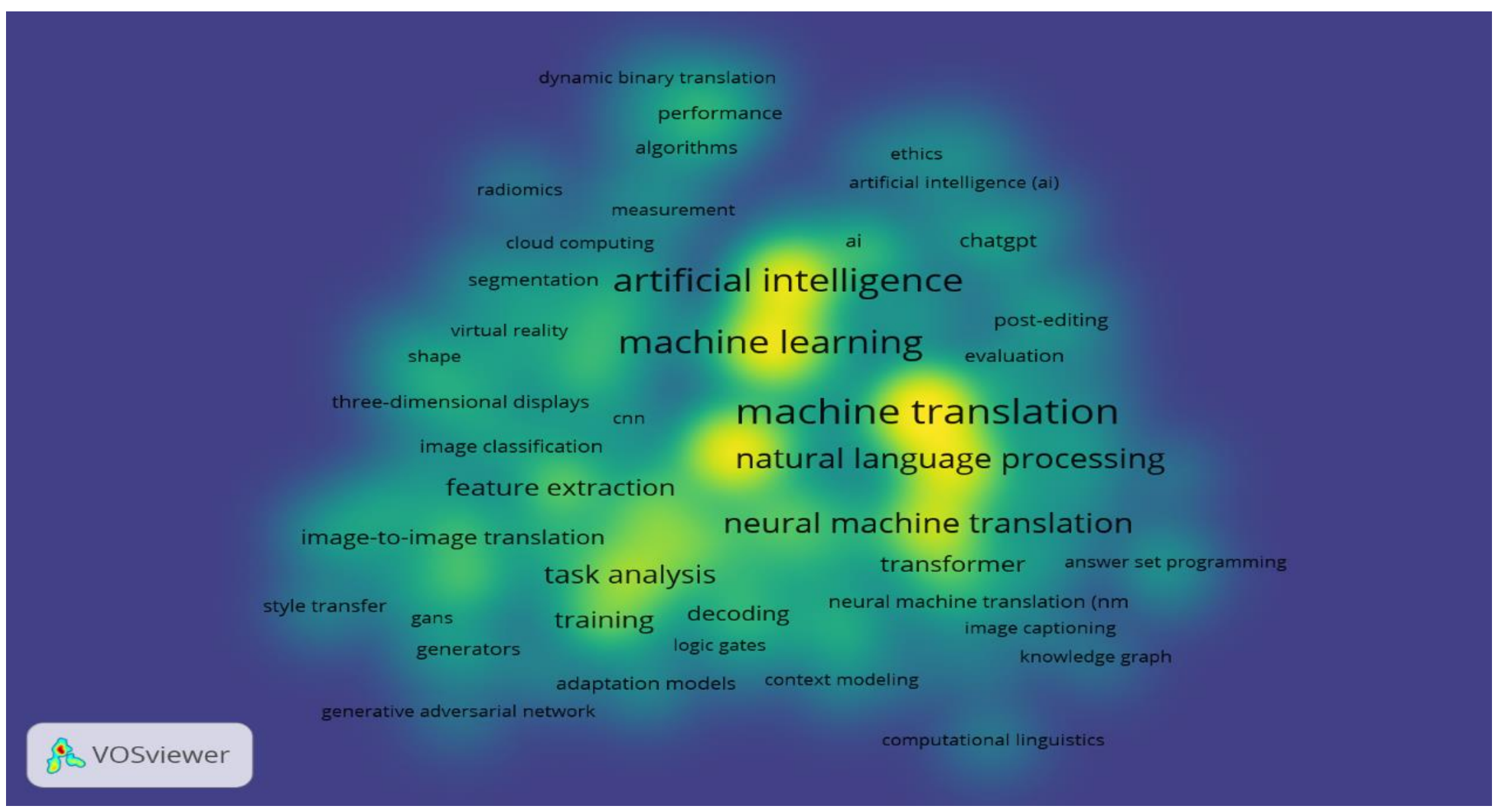

***Fig 6: Density view of Author's Keywords***

As Fig 6 displays, the Density view of author's keywords seems to mirror both the cluster and subject categories analyses. Put differently, the Density view of the author's keywords covers all the terms occurring in these two categories. It also reflects and covers the terms occurring in the categories *Nodes by bursts* and *Nodes by centrality* as we will see shortly.

### 4.2.3. Nodes

Table 6 presents the most important nodes by Bursts in ACTI production during forty-five years. *Statistical machine translation* is the top ranked item by Bursts in Cluster 1, with Bursts of 56.52, beginning in 1999, and highlighting ACTI developments during 2003-2019. The 2nd is *Information retrieval,* Cluster 0, with Bursts of 17.09 marking a stage of ACTI, and ensuring that information retrieval precedes SMT, which perhaps indicates a stage of developing programs to be used in SMT. The 3rd is *Translation quality* in Cluster 1, with Bursts of 16.25, accelerating the need of good translation and pinpointing the bad quality of MT output. The 4th is *Target language* in Cluster 8, with Bursts of 15.38. This keyword represents a search area in ACTI, manifesting that TL should be paid much attention to. The 5th is *New approach,* Cluster 1, with Bursts of 15.13, indicating perhaps a change towards a new approach in MT industry. The 6th is *Support vector machine* in Cluster 7, with Bursts of 14.74. The 7th is *Machine translation system*, Cluster 1, with Bursts of 13.02. The 8th is *Cross-language information retrieval* in Cluster 0, with Bursts of 12.40. The 9th is *Parallel corpora* in Cluster 1, with bursts of 11.52. The 10th is *transformers* in Cluster 2, with Bursts of 11.00. Notice that almost all top 10 nodes by Bursts iterate in our previous analyses. This gives us a room to postulate that our analysis is reliable and valid.

**Table 6: Top nodes by Bursts**

| Bursts | Node Name | Cluster ID |
|---|---|---|
| 56.52 | statistical machine translation | 1 |
| 17.09 | information retrieval | 0 |
| 16.25 | translation quality | 1 |
| 15.38 | target language | 8 |
| 15.13 | artificial intelligence | 1 |
| 14.74 | support vector machine | 7 |
| 13.02 | machine translation system | 1 |
| 12.40 | cross-language information retrieval | 0 |
| 11.52 | parallel corpora | 1 |
| 11.00 | transformers | 2 |

### 4.3. Top nodes by Centrality

In scientometric analysis, Centrality reflects the likelihood of an arbitrary shortest path in the network. It is also regarded as a position *between* two large sub-networks (Chen, 2006; Ballouket al., 2024). Table 7 showcases the top ranked nodes by Centrality in ACTI production during forty-five years. The 1st is *Machine translation* in Cluster 0, with Centrality of 0.27. The 2nd one is *Novel approach* in Cluster 13, with Centrality of 0.07. The 3rd is *Artificial intelligence* in Cluster 3, with Centrality of 0.07. The three last node by Centrality are *Natural language processing* in Cluster 0, *ChatGPT* in Cluster 14, and *Support vector machine* in Cluster 7. ChatGPT has been reported to perform competitively better than other nonneural tools and apps like Google Translate, specifically, with GPT-4 (Jiao et al., 2023).

**Table 7: Top Nodes by Centrality**

| Centrality | Node Name | Cluster ID |
|---|---|---|
| 0.27 | machine translation | 0 |
| 0.07 | novel approach | 13 |
| 0.07 | artificial intelligence | 3 |
| 0.06 | machine learning | 0 |
| 0.06 | logic programming | 6 |
| 0.05 | statistical machine translation | 1 |
| 0.05 | target language | 8 |
| 0.05 | natural language processing | 0 |
| 0.05 | chatgpt | 14 |
| 0.05 | support vector machine | 7 |

Again, notice that almost all top nodes by Centrality are iterated from previous analyses, be they related to Cluster, Subject categories, Keywords or top nodes by Bursts.

### 4.4. Top references by Citation Bursts

As Fig 7 displays, DCA of the references extracted from CiteSpace 6.3.1R depends on the strongest Bursts. In Fig 8, Vaswani et al. (2017) is the article whose citation Bursts is 62.1, lasting from 2021 to 2022. The second article with strongest citation Bursts is Bahdanau et al. (2016) with citation Bursts of 35.6; its strongest citation was during the period between 2017 and 2021. The Bursts then goes down reaching its lowest status with Simonyan et al. (2015), which ranks 12$^{th}$, and whose strength of Bursts is 13.94 lasting from 2018 to 2020. Additionally, we utilized DCA analysis due to its ability to pinpoint the trending issues and hotspots of ACTI during forty-five years (cf. Chen, 2017, Chen et al., 2010). In fact, DCA analysis is more reliable, giving us essential and unique information to specify the trending issues and hotspots in ACTI research. DCA is more influential than Author Co-citation Analysis (ACA). It is more prolific than ACA, unveiling more specific information and patterns than ACA. Further, it is easy to identify a cited publication in a DCA network more than ACA. The former also allows for more clarity and less ambiguity than the latter.

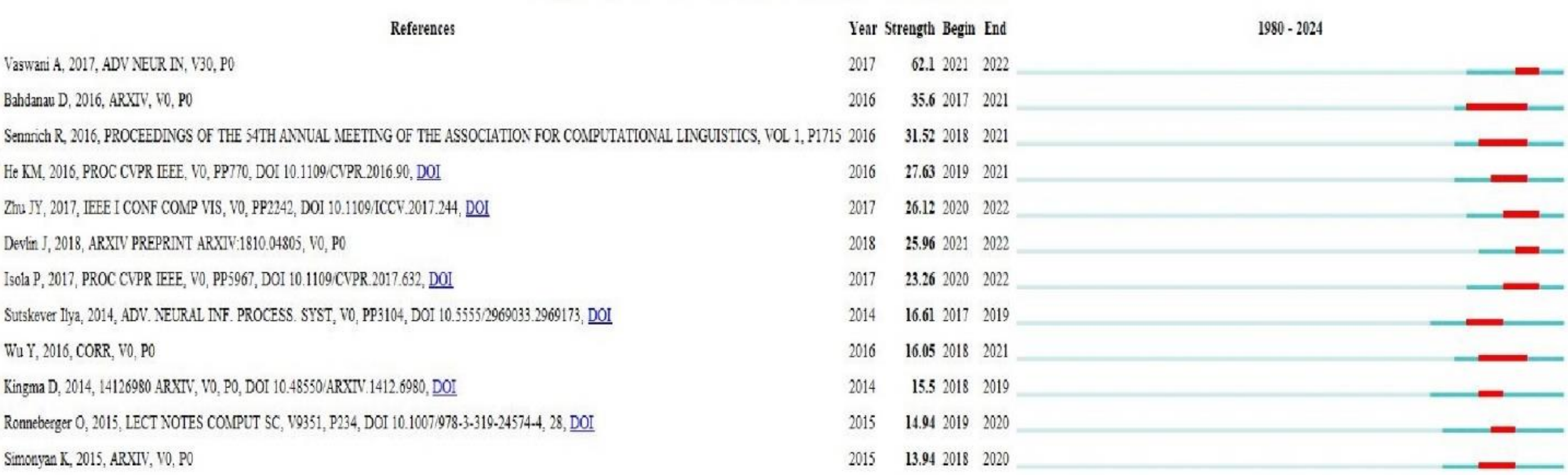

**Top 12 References with the Strongest Citation Bursts**

| References | Year | Strength | Begin | End | 1980 - 2024 |
|---|---|---|---|---|---|
| Vaswani A, 2017, ADV NEUR IN, V30, P0 | 2017 | 62.1 | 2021 | 2022 | |
| Bahdanau D, 2016, ARXIV, V0, P0 | 2016 | 35.6 | 2017 | 2021 | |
| Sennrich R, 2016, PROCEEDINGS OF THE 54TH ANNUAL MEETING OF THE ASSOCIATION FOR COMPUTATIONAL LINGUISTICS, VOL 1, P1715 | 2016 | 31.52 | 2018 | 2021 | |
| He KM, 2016, PROC CVPR IEEE, V0, PP770, DOI 10.1109/CVPR.2016.90, DOI | 2016 | 27.63 | 2019 | 2021 | |
| Zhu JY, 2017, IEEE I CONF COMP VIS, V0, PP2242, DOI 10.1109/ICCV.2017.244, DOI | 2017 | 26.12 | 2020 | 2022 | |
| Devlin J, 2018, ARXIV PREPRINT ARXIV:1810.04805, V0, P0 | 2018 | 25.96 | 2021 | 2022 | |
| Isola P, 2017, PROC CVPR IEEE, V0, PP5967, DOI 10.1109/CVPR.2017.632, DOI | 2017 | 23.26 | 2020 | 2022 | |
| Sutskever Ilya, 2014, ADV. NEURAL INF. PROCESS. SYST, V0, PP3104, DOI 10.5555/2969033.2969173, DOI | 2014 | 16.61 | 2017 | 2019 | |
| Wu Y, 2016, CORR, V0, P0 | 2016 | 16.05 | 2018 | 2021 | |
| Kingma D, 2014, 14126980 ARXIV, V0, P0, DOI 10.48550/ARXIV.1412.6980, DOI | 2014 | 15.5 | 2018 | 2019 | |
| Ronneberger O, 2015, LECT NOTES COMPUT SC, V9351, P234, DOI 10.1007/978-3-319-24574-4_28, DOI | 2015 | 14.94 | 2019 | 2020 | |
| Simonyan K, 2015, ARXIV, V0, P0 | 2015 | 13.94 | 2018 | 2020 | |

*Fig 7: Top 12 References by Citation Bursts*

### 4.5. Top research centers by Citation

Fig 8 portrays the top twelve Research Centers contributing to ACTI production in the world, sorted by number of citations. MIT ranks 1$^{st}$, with 3421 citations and 85 publications. Sandford University ranks 2$^{nd}$ with 2745 citations and 53 publications. The 3$^{rd}$ and 4$^{th}$ ranks are retained by University of California, Berkeley (2427 citations, 30 publications) and Carnegie Mellon University (2393 citations and 51 publications), respectively. The 5$^{th}$ research center is Microsoft Corporation with 1609 citations and 8 publications. The last two research centers are King Saud University and Nanyang Technological University with 1188 citations and 23 publications, and 1180 citations and 18 publications, respectively. Almost all research centers are historically deep rooted in scientific research, specifically computer science and technology, CL and NLP. They belong to USA, Germany, Canada, China, Singapore and one belongs to Arab world, namely Saud Arabia.

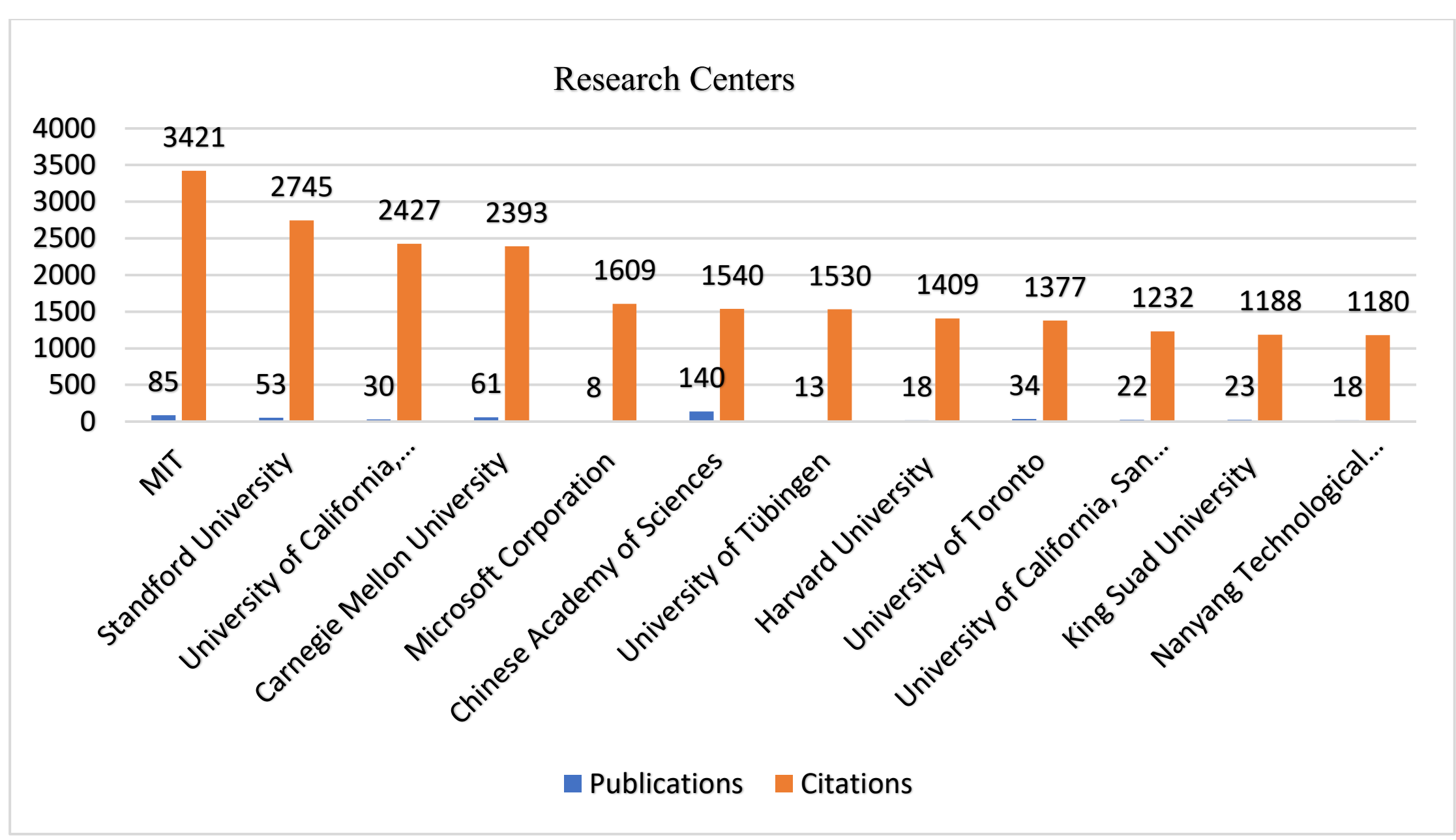

*Fig 8: Top 12 key research centers in ACTI*

### 4.6. Thematic analysis

In this section, we focus on reviewing 18 articles related to ACTI, carefully and purposefully chosen from the articles involved in this study. The review is conducted in terms of author, purpose/aim, methodology/approach, findings and contribution to ACTI future directions. Our rationale of focusing on these aspects is based on Khosravi et al.'s (2024) methodology. The reviewed articles are the most recent publications in our corpus. If there were more than one article in/about a certain topic of ACTI, we consider the most recent one(s) in our data.

The articles reviewed in Table 8 encompass several themes, we have set our criteria to eight categories: SMT, NMT, Post-editing, Linguistics, Language-specific, Hybrid, ChatGPT, and Transformer. SMT include two studies, each of which applies AI technology to enhance MT. NMT categories involves four studies. Post-editing category reviews a study on involving human post-editing. Linguistics include two studies, one involving syntax and the other semantics. Language-specific comprises three studies. In this category, we focused on ACTI studies centered around specific language pairs such as English and Chinese. Hybrid category has four studies, which we focus on involving both SMT and NMT models. ChatGPT category involves two studies, examining the usefulness of ChatGPT in NMT. Finally, Transformer includes one study in which the authors develop a transformer model for MT. These 18 articles represent ACTI spectrum during forty-five years. Note that there is some sort of overlap in the classification, but we tried to classify the selected articles including each article in the category it fits more.

The thematic review of the 18 articles allows for more 'focused insights and spans various themes such as statistical MT, neural MT, hybrid models, and generative tools including ChatGPT and Transformers'. It also results in other themes such as the role of post-editing in MT, contribution of linguistics to MT including works on CL, and NLP. One of the most interesting results in this juncture is that the review highlights MT works on specific languages such as applying AI to Chinese MT, incorporating CL, and exploring the strengths and limitations of an AI-based English-Chinese translation of literary texts. Other languages include the Mixtec language from which MT translates into the Spanish language, via introducing a novel method for collecting and translating texts from this language. The Mixtec language is an indigenous language spoken by people of southern Mexico. Thus, that involving such indigenous languages in MT with the help of AI is a remarkable achievement.

I have also adopted thematic saturation (see e.g. Guest et al., 2020) so that the selected articles reveal recurring patterns and insights across the targeted criteria. To ensure thematic depth and relevance, we employed purposeful sampling (see e.g. Palinkas et al., 2015), a widely accepted approach in qualitative research. The selection was also based on the alignment with the eight core thematic categories that reflect the most significant areas of AI's contribution to ACTI production, namely BMT, SMT, NMT, post-editing effects on MT outputs, Linguistic influence on translation industry, represented by NLP approaches, Language-specific applications. Another crucial area in our representative data concerns hybrid MT models. Such data were involved to pinpoint the fact that single models may not work properly on some data of languages such as low-resource languages, hence hybrid models may resolve these problems. The same thing, albeit different in focus, could be said about other criteria such as Transformer-based architectures and generative AI tools such as ChatGPT. Given the purposeful selection method, the articles involved were included if they made a meaningful contribution to at least one of these thematic areas. Our aim was to capture the breadth of methodological, technological, and linguistic diversity within ACTI production. Thus, the 18 purposefully selected articles and the analysis employed result in an in-depth, qualitative exploration of key themes emerging from ACTI production, as presented in Table (8).

**Table 8: Thematic review of the literature on ACTI**

| No | Author(s) + year | Aim/purpose | Approach/methods | Findings | Contribution to ACTI future directions |
|---|---|---|---|---|---|
| **SMT** | | | | | |
| 1 | Formiga et al. (2015) | Improving a SMT system by incorporating human post-editing. | Mixed methodology evaluating a real-world dataset collected from Reverso.net | The proposed AI model is robust in automatically Web-crawled parallel corpora.<br>It enhanced SMT translation | The proposed system can be applied to Arabic corpora like [Falak](#) |
| 2 | Li et al. (2023) | Comparing conventional SMT with an AI-based translation model to produce high-quality translations. | comparative/experimental methodology | conventional SMT achieves 4.9667fluency while AI-based translation model 6.6333. | This AI model can be applied to Arabic mismatchings of L dialects, hence improving ACTI in Arabic context. |
| NMT | | | | | |
| 3 | Lo (2023) | Improving learners' vocabulary in EFL classrooms by utilizing Neural machine translation: | a quasi-experimental approach | there was no big change in higher proficiency learners' achievement in vocabulary retention.<br>- NMT model proposed helps lower proficiency learners' achievement in immediate vocabulary retention. | The proposed model can be applied in teaching ACTI in Arabic-speaking classrooms, specifically with low proficiency learners. |
| 4 | Belinkov et al. (2020) | Analyze the representations learned by NMT models at various levels of granularity and evaluate their quality through relevant extrinsic properties | Experimental methodology focusing on training NMT data | - Word morphology features are simpler than those of non-local syntactic and semantic dependencies<br>- Representations learned using full words are more informed than those learned using subword parts.<br>- Multilingual LLMs are richer than bilingual ones. | Multilingual LLMs are more valid than bilingual ones, hence enriching ACTI field. |
| 5 | Huang et al. (2023) | introducing a multi- NMT model | Experimental methodology (focusing on generating target sentences by NMT's proposed formulae) | The proposed AI model significantly improved translation performance on a strong baseline. | the proposed model can be applied to Arabic dialects perhaps along with Baniata & Kang's (2024) model. |
| Post-editing | | | | | |
| 6 | Sánchez-Gijón et al. (2019) | Addressing differences between NMT post-editing and translation with the aid of a translation memory (TM). | Empirical methodology (Eight professional English–Spanish translators took part in this test. | - NMT post-editing involves less editing than TM segments.<br>- translators positively thinking performed faster than those thinking of it negatively. | the proposed model can be extended and applied to Arabic English NMT, hence enhancing ACTI |
| Linguistics | | | | | |
| 7 | Chen et al. (2019) | proposing a new neural network with syntax-based convolutional architecture to learn structural syntax information in translation contexts.<br>- improving translation output. | Experimental methodology | the proposed model can achieve a substantial and significant improvement over several baseline systems. | Given that the proposed model is related to improvements of NMT, it could be applied to ACTI involving Arabic dialects. |
| 8 | Song et al. (2019) | proposing a semantic NMT model for MT incorporating abstract meaning representation (AMR) | Experimental methodology | the proposed model improves a strong attention-based sequence-to-sequence NMT with reference to English and German | The proposed model can be further applied to similar connects in MT. |
| Language-specific | | | | | |
| 9 | Feng et al. (2023) | Applying AI to Chinese MT incorporating CL | Experimental approach | The proposed model of MT improves Chinese translation in both performance and accuracy | The proposed model could be applied to Arabic translation to English. Arab computational linguists and AI specialists could develop ArabNet corpus and develop Translation Algorithms |
| 10 | Hu & Li (2023) | exploring the strengths and limitations of an AI-based English-Chinese translation of literary texts. | corpus-based approach | - The proposed AI model performs well, resulting in 80% accuracy in translating Shakespeare's plays *Coriolanus* and *The Merchant of Venice*<br>- It exhibits some sort of creativity. | The proposed model could be extended to translating famous Arabic literary like those written by Mahmoud Alaqad, Taha Hussain, among others. |
| 11 | Santiago-Benito et al. (2024) | introduces a novel method for collecting and translating texts from the Mixtec to the Spanish language | a mixed approach | achieved a bilingual evaluation understudy (BLEU) score of 95.66 for Mixtec-to-Spanish translation and 99.87 for Spanish-to-Mixtec translation. | trained automatic translation models based on recurrent neural networks, bidirectional recurrent neural networks, and Transformers |

| Hybrid | | | | | |
|---|---|---|---|---|---|
| 12 | Jung et al. (2024) | proposes combining Google Translate & ChatGPT) and ANNs concerns of human scoring | comparative approach (multilingual student responses from eight countries and six different languages were recruited as participants) | Automated scoring displayed comparable performance to human scoring, especially when the ANNs were trained and tested on ChatGPT-translated responses | highlights that automated scoring integrated with the recent machine translation holds great promise for consistent and resource-efficient scoring in ILSAs |
| 13 | Tosun (2024) | Proposing a Turkish-English model for detecting accuracy in MT | Experimental methodology | Late bilinguals more effectively detect MT accuracy than their early bilingual counterparts. | Concerning the preference for MT, age of acquisition and the accuracy detection of non-firsthand sentence translations emerged as significant predictors. |
| 14 | Dugonik et al. (2023) | proposing a hybrid machine translation (HMT) of both NMT and SMT. | Experimental approach | The proposed HMT system:<br>- boosted the BLEU score, with an increase of 1.5 points and 10.9 points for both translation directions.<br>-contributed to Slovenian–English translation field | The model could be applied to other languages leveraging the multilingual language model proposed. |
| 15 | Wang X. et al. (2018) | proposing an AI model for incorporating SMT system into NMT to alleviate above word-level limitations | experimental methodology (Chinese-to-English and English-to-German translations) | The proposed model performs better than NMT and SMT each alone. | Given the high peculiarities of CL, SA, and MAD, the proposed model could be applied to these Arabic verities and other languages including English. |
| ChatGPT | | | | | |
| 16 | Peng et al. (2023) | Introducing ChatGPT as the best AI model for MT | Experimental methodology involving English-Centric Language Pairs | The chain-of- thought prompt was found to be powerful leading to word-by-word translation, which brought invaluable translation degradation. | Enhancing ChatGPT functionality. |
| 17 | Son and Kim (2023) | Comparing ChatGPT translation with that of Google Translate | Comparative methodology | - ChatGPT produces more reliable translation than CATs.<br>- Translation into English is easier than from it. | The proposed token-based MT system can be applied to translating Arabic culture-based expressions into English, and vice versa. |
| Transformers | | | | | |
| 18 | Baniata et al. (2021) | Introducing a transformer model to translate Arabic dialects. | Experimental methodology | - The proposed model enhances and accelerates the importance of various NLP tasks.<br>- It enriches NMT providing an AI-based model for Arabic dialects. | the proposed model has been applied to Arabic dialects, hence contributing to solving the problems encountered by MT in dealing with free word order languages. |

## 5. Discussion

This paper synthesizes ACTI intellectual landscape over forty-five years from 1980-2024, considerably a long timeframe aiming to map it out and pinpoint the trending issues in this very important area of human knowledge. We have focused on providing analyses of ACTI knowledge landscape and trending issues, represented by clusters, subjects categories, keywords, nodes by bursts and centrality, as scientometric indicators to characterize, visualize and embody ACTI prodiction in this period. Three influential sources the articles were retrieved from are WoS, Lens, and Scopus.

To elaborate on a point alluded to so far, Cluster analysis in the scientometric analysis gives us profound insights into several factors, the most important of which are: i) a holistic picture of ACTI intellectual landscape during forty-five years, ii) trending issues in ACTI production, and iii) its hotspots. As for (i), CiteSpace identified 471 research trends constituting ACTI knowledge landscape, we discussed only 12 clusters (Table 4). As for (ii), trending issues refer to those clusters representing a high number of publications, CiteSpace dubbed these as “Sizes”. Concerning (iii), hotspots are articles that are constituted recently. Thus, in 1980-2024 timeframe the top 12 clusters beginning with *Small parallel corpora* and ending in *Deep learning models* constitute

ACTI knowledge landscape. Clusters such as those having the sizes 125, 117, and 100 could be termed as trending issues, and clusters evolving around 2021-2023 could be called hotspots in ACTI over the study timeframe. We will discuss these and their relevance to ACTI developments, focusing on (ii) trending issues and (iii) hotspots, each in turn.

ACTI trending issue *Generative adversarial network* (Cluster 0), comes to play around 2017, having the highest number of publications, i.e. 125 articles, and a Silhouette value of 0.96. Generative adversarial network is a type of LLM, whose main idea comes from game theory of gain and loss. It is an LLM combined with two neural networks for adversarial training, recently introduced into AI field, specifically deep language learning to produce highly accurate images, utilizing deep learning methods in face recognition (Long & Zhang, 2023; Tavakkoli et al., 2020). The importance of this ACTI trend ensues from utilizing GAN in MT, though it firstly was introduced in face recognition in military field and forensic and legal spheres. It has been employed in natural language processing including MT, thus generating language texts/sentences to overcome MT translation problems (Ahn et al., 2021).

*Low-resource language* is another ACTI trend retaining the 2nd rank, i.e. Cluster 1 with 117 research articles and a Silhouette value of 0.883. Low-resource languages refer to languages having less data for AI models to process and/or analyze including Irish, Hindi, Bengali, Amharic, Galician, and Basque. These languages create problems for LLMs, because they do not provide rich data for language processing in machine translation processes, for instance, in the same way high-source languages such as English, Arabic, French, and German do. Low-resource languages thus need certain types of deep learning models capable of processing their poor data. A number of ACTI scholars have attempted to build such models. For example, Goyle et al. (2023) build an NMT model capable of processing *low-resource language* data. Their model was built on mBART, which seems to achieve promising results helping in amending these languages' problem space. In this aspect, Pang et al. (2025) revisit six core challenges such as domain mismatch, limited parallel data, rare word prediction, long sentence translation, attention model as word alignment, and sub-optimal beam search. Their findings indicate that LLMs have made significant strides in mitigating some of these issues.

*Statistical machine translation* is another trending issue in 1980-2024, labelled as Cluster 3 (Table 4). It was established around 2011, with 94 articles and a Silhouette value of 0.974. SMT has a long history, as one of three stages of MT. It has come after *Example-based machine translation* and developed from it. The first idea of SMT is traced to Brown et al.'s (1990) proposal. In SMT, machines, precisely computers, are trained on a large amount of data to learn translation rules (Wang H. et al., 2022). Since then, SMT has developed in several aspects including different models (Brown et al., 1993), a syntax-based model (Yamada & Knight, 2001), alignment template approach (Och & Ney, 2004), shallow linguistic knowledge (Hwang et al., 2007), hierarchical phrase alignment (Watanabe et al., 2007), swarm-inspired re-ranker system (Farzi & Hesham, 2015), SomAgent models (López et al., 2011), syntax-based reordering model (Khalilov & Fonollosa, 2011), and incorporating human post-editing (Formiga et al., 2015). However, there have also arisen some studies combining multi-models (Sachdeva et al., 2014), because it was not easy for a single model to meet several translation requirements, which enhances SMT performance (see also Dugonik et al., 2023). The use of SMT has been perhaps limited when NMT comes to play, which witnesses substantial developments, specifically when employing neural algorithms, though hybrid models have been proposed. For example, Taheri et al. (2024) introduce a hybrid data augmentation technique aiming at improving Aspect-Based Sentiment Analysis (ABSA) performance without the need for additional labeled data. This ABSA model focuses on identifying specific aspects within a sentence and determining the sentiment associated with each aspect.

As for ACTI hotspots during forty-five years, three clusters, namely *Arabic dialects, Multi-modal neural machine translation* and *Deep learning models* are a case-in point here. For example, *Arabic dialects* as a hotspot in ACTI research evolves around 2020, with 16 publications in 1980-2024 timeframe. The entrance of Arabic language/dialect is very significant due to the fact that Arabic has three major categories/dialects: Classical Arabic (CLA), Standard Arabic (SA) and Modern Arabic dialects (MADs). CLA is the language of the Holy Qur'an, it is used in religion-based contexts such as mosques, Islamic teachings, while SA is the official language of the Arab States. It is used in such context as press, TV, education, official communications. Finally, MADs are mother tongues of all Arabs. These dialects are used in day-to-day affairs, homes and streets. Undoubtedly, these three varieties of Arabic impose severe problems for MT, be it SMT or NMT. Thus, there are ample studies examining the efficiency of MT for Arabic dialects, accelerating Arabic's peculiarities and properties. For example, Baniata et al. (2021) have introduced a pioneering transformer model that utilizes sub-word units for translating Arabic dialects, holding that Arabic dialects constitute considerable difficulty for NMT as they exbibit free word order and dialectical lexes.

As for other analyses such as Subject categories, Keywords, Top nodes by Bursts and by Centrality, it is clear that they are consistent with Cluster analyses. Put differently, we notice that almost all the clusters in Table 4 iterate in these analyses, which accelerates the strength, reliability and validity of our analysis. A substantial finding from Table 4 concerns the size of the contribution of AI to translation industry. The fact that *Neural machine translation, Generative adversarial network* and *Low-resource language* have the largest number of articles, viz., 100, 125 and 117, respectively, compared to *Small parallel corpora,* which emerges around 2004, unveils that AI contribution to translation industry passes through gradual stages, and that the more AI develops the more it contributes to MT.

Regarding 2015-2024 timeframe (Table 5), ACTI intellectual production undergoes considerable changes. Comparing Table 4 to Table 5, we notice that several trends in ACTI were missing. These include five clusters: *Small parallel corpora, Wide-coverage multilingual semantic network, Large language model, Arabic dialects,* and *Deep learning model*s. The other seven clusters remain in 2015-2024 timeframe, but with several changes in cluster ID, Size, Silhouette and year of emergence. For example, in 1980-2024 timeframe *Generative adversarial network* (Cluster 0), emerging in 2017, has 125 articles and a Silhouette value of 0.96. However, in 2015-2024 timeframe it reemerges around 2019, having 138 articles and a Silhouette value of 0.864. Another example is that around 2018, Cluster 1 *Low-resource language* has 117 articles in 1980-2024, which increases by 8 articles in 2015-2024 timeframe. Further, there are five clusters that emerge in 2015-2024 timeframe. These five clusters are: *Sequence-level learning, Convolution-enhanced evolving attention network, Document-level neural machine translation, Multi-modal neural machine translation,* and *Function graph call context*. These newly emerging ACTI trends could be thought of as hotspots in 2015-2024 timeframe.

In both timeframes, we clearly see what AI contributes to translation industry, bringing about several and varied developments in MT, starting from RMT passing through SMT and NMT, and now hybrid models, e.g. both SMT and NMT. These hybrid models can handle several emerging issues like low-resource languages including Bengali, Amharic, and Galician, multidialectal languages like Arabic, among many other issues. There are also attempts now to incorporate cultural and religious nuances in AI training data. There are also efforts to develop AI-driven translation models for indigenous languages such as the Mixtec language (see e.g. Santiago-Benito et al., 2024).

Turning to thematic analysis, we conducted the thematic review considering determining the themes of ACTI production and to what extent AI contributes to translation industry. we will just discuss one article from each category. It should be noted here that the category *Contribution to ACTI future directions* in Table 8 was meant to provide possible solutions for problems encountered by NMT in dealing with low-resource, multi-dialectical and free word order languages.

Article 1 by Formiga et al. (2015) introduces a mechanism for integrating user feedback into SMT systems, effectively turning them into dynamic learning systems. It represents *SMT* category, giving a vivid clue of how static models have shifted to adaptive ones, where AI systems evolve based on real-world interactions. The authors focus on real-time retraining, enhancing the idea of "continual learning", as a growing area in AI study. As for their contribution to ACTI future direction, the proposed system can be applied to Arabic corpora like Falak.

Article 3 by Lo (2023) represents *NMT* category. It provides a mechanism for improving learners' vocabulary in EFL classrooms by utilizing Neural machine translation. The NMT model proposed helps lower proficiency learner's achievement in immediate vocabulary retention. The proposed model can be applied in teaching translation in Arabic-speaking classrooms, specifically with low proficiency learners.

Article 6 by Sánchez-Gijón et al. (2019) belongs to *post-editing* category. It addresses differences between NMT post-editing and translation with the aid of a translation memory (TM). NMT post-editing involves less editing than TM segments. This area enriches AI contribution to translation industry. The proposed model can be extended and applied to Arabic English NMT, hence enhancing neural MT.

Article 7 by Chen et al. (2018) is within *linguistics* category. Our aim to include this article stems from our intention to assess how linguistics amalgamated with AI contributes to translation industry. This paper proposes a new neural network with syntax-based convolutional architecture to learn structural syntax information in translation contexts. The findings indicate that utilizing this model improves translation output. The proposed model is related to improvements of NMT, thus it could be applied to MT involving Arabic dialects.

Article 9 by Feng et al. (2023) belongs to *language-specific* category. The paper focusses on applying AI to Chinese MT incorporating CL Experimental approach. The proposed model of MT improves Chinese translation in both performance and accuracy. The proposed model could be applied to translating Arabic into English. Arab

| The proposed model can be applied to multi-language corpora. |
|---|
| - The proposed model could be applied to social media outlets including WhatsApp, Facebook, across 100 languages.<br>- It could also applied to Arabic modern multi-dialectical varieties. |

computational linguists and AI specialists could develop *ArabNet* corpus and develop suitable Translation Algorithms.

Article 15 by Wang X. et al. (2018) is within the category *hybrid*, proposing an AI model for incorporating SMT system into NMT to alleviate above word-level limitations. It involves Chinese-to-English and English-to-German translations. The proposed model performs better than NMT and SMT each alone. Given the high peculiarities of CL, SA, and MADs, the proposed model could be applied to these Arabic verities and other similar languages.

Article 17 by Son and Kim (2023) represents the *ChatGPT* category. The authors compare ChatGPT translation with that of Google Translate. ChatGPT produces more reliable translation than nonneural tools. It is found that translation into English is easier than from it. The proposed token-based MT system can be applied to translating Arabic culture-based expressions into English, and vice versa.

Article 18 by Baniata et al. (2021) lies within the category *Transformer*. The authors introduce a transformer model to translate Arabic dialects. It is found that the proposed model enhances and accelerates the importance of various NLP tasks. It enriches NMT providing an AI-based model for Arabic dialects. It has been applied to Arabic dialects, hence somehow contributing to solving the problems encountered by MT in dealing with free word order languages.

Finally, two studies published in 2025 propose multi-lingual and multi-model LLMs and multi-purposes architecture. These two studies may be considered a thematic category in our study added to our eight categories. (These two articles were recommended by a *DAI* reviewer, many thanks to him/her). The first is conducted by SEAMLESS Communication Team (2025), proposing a unified multi-lingual and multi-modal machine translation system, capable of handling both speech and text across up to 100 languages. This innovative model integrates multiple translation tasks: speech-to-speech, speech-to-text, text-to-speech, and text-to-text, into a single architecture. The second study is proposed by Pang et al. (2025), which revisits six core challenges: domain mismatch, limited parallel data, rare word prediction, long sentence translation, attention model as word alignment, and sub-optimal beam search. Their findings indicate that LLMs have made significant strides in mitigating some of these issues, but there are some challenges that still persist including LLMs' inability to function properly in low-resource languages. As for how this category can contribute to ACTI future frontiers, SEAMLESS Communication Team's proposed model can be applied to social media outlets including WhatsApp, Facebook, across 100 languages. It could also be applied to Arabic modern dialects' speech<->to<->text, given that these Arabic dialects are multi-dialectical, but they are also sometimes unintelligible. Pang et al.'s (2025) model can be applied to large corpora, hence overcoming 5 n-gram limitations, a problem encountered in AI models' prediction (see e.g. Gulordava et al., 2018). This added category clearly represents AI specialists' continuous efforts to better LLMs' functionality.

## 6. Conclusions and looking forward

There seems to be a parallel correspondence between AI and MT developments, the development of the former entails the improvement of the latter. Through history spanning over a century or so, MT has undergone several developments, beginning with RMT, SMT and ending in NMT and transformers, LLMs and hybrid SMT, and NMT models. The study provides comprehensive scientometric and thematic analyses of ACTI research for a considerably long time spanning between 1980 and 2024. It involves 9836 articles collected from three different but reliable sources, viz., WoS, Scopus, and Lens. The scientometric analysis involves Cluster, Subject Categories, Keywords, Bursts, Centrality and Research Centers. The thematic analysis focuses on reviewing 18 articles, selected purposefully from the top citing and cited articles, centering on author, purpose, approach, findings, and contribution to ACTI future directions, and coming up with several themes.

Several conclusions could be drawn from our study: i) the more AI develops, the more translation industry improves, resulting in RMT and SMT whose translation output was not that satisfactory. This "unsatisfactoriness" could be ascribed to many factors, some related to the undeveloped models used, and some related to typological parametrizations between languages, ii) with the advent of NNAs and DLLMs utilized in MT, MT industry develops considerably, specifically with launching large LLMs such as ChatGPT, and utilizing several transformers (see e.g. Santiago-Benito et al., 2024). These AI models have revolutionized almost every aspect of life including academics, education, engineering, scientific research, and MT industry is no exception, and iii) despite all these developments, MT field still needs rigorous projects to better MT translation outputs. NMT along with current AI models need to develop further to handle many problems. These problems include those related to low-resource languages such as Hindi, Bengali, Amharic, and Irish. Due to the poor linguistic data these languages have, models developed for them need further research to process their low sources (Goyle et al., 2023; Pang et al., 2025).

However, there are several areas of challenge for AI-driven translation. One such area presenting a real challenge concerns multi-dialectical and free word order languages like Arabic. Due to the diglossic nature of Arabic (see e.g. Ferguson, 1959), we need an AI model capable of handling, processing and computing H and L varieties of Arabic. Arabic, be it CLA, SA, or MAD, has a variety of word orders such as VSO, SVO, SOV, OSV, OVS (Shormani, 2015, 2017, 2024a, b &c), so we need NMT models that can identify the subject, verb and object in the sentence to translate them correctly (see also Baniata et al., 2021). A further problem encountered by NMT is with regard to register, specifically cultural and religious registers. As for culture, ChatGPT, though the most powerful AI model, cannot translate an English proverb like *The devil is beating his grandma* (a proverb said when rain falls and sun shines) into Arabic, as proverbs involve arbitrariness and idiosyncrasies, and are embedded within culture (cf. Shormani, 2020). As for religious texts, ChatGPT, for instance, cannot translate such an Arabic religious text as بحجر الله! 'bi-ħajr ʔallah!' (=For the sake of God) (cf. Qarabesh et al., 2023). Thus, what AI translation programmers and developers should do in this respect is train LLMs like ChatGPT on massive (internet) data that include religious and cultural data (here Arabic and English) to better its translation performance.

Additionally, the severe challenges AI, and AI specialists encounter, which also make common man afraid of AI future, are ethics and biases. AI specialists should ensure that LLMs are ethically considerate, and bias-free. As for ethics, for example, LLMs should provide equitable outcomes across different demographic groups (Mutashar, 2024). These outcomes will lessen the challenges arising when these models inadvertently perpetuate existing societal groups. Thus, if these outcomes are adhered by, we can ensure fair treatment of certain groups. LLMs are considered "black boxes" making it difficult to understand how decisions are made. Ethical LLMs should be interpretable so users can trust and verify their information. An important aspect is or concerns privacy and data protection. For training, LLMs depend on massive amounts of (internet) data, which can include sensitive personal information (Feng et al., 2023). This goes against personal privacy. In this respect, AI specialists should ensure compliance with privacy laws and at least minimize data misuse or plagiarism (see e.g. Baniata et al., 2021; Feng et al., 2023; Lo, 2023).

As for AI biases in LLMs, using AI-based tools may lead to destructive outcomes. For instance, a recruitment algorithm trained on male-dominated job applications may prefer male applicants. The design of algorithms can reveal bias if certain groups are not adequately represented or accounted for (see also Lion et al., 2024, and references there). LLMS that incorporate user feedback, as discussed in Formiga et al. (2015), can inadvertently reinforce biases present in user behavior. For example, if most users consistently prefer certain translations or outcomes, their preferences may be overlooked. LLMs should also reflect equal representations of all societal groups. This is due to the fact that some groups including disabled people, or any ethnic group may be underrepresented or misrepresented in the training data, leading to bias against these individuals. Additionally, there is some sort of bias in linguistics. For instance, specific language(s) are used as training (data) languages, neglecting other languages. LLMs seem to be predominantly based on English. This is very clearly shown in our study, as represented by low-resource and free word order languages, for example.

Due to the problematic issues encountered in AI-based translation industry, the following actionable practices for integrating AI into human translation workflows could be highlighted: i) initial drafts. AI tools can assess text complexity, terminology consistency, and potential challenges before human translators begin their work, allowing for better preparation and resource allocation. LLMs such as ChatGPT, DeepSeek, and DeepL can serve as drafting tools to generate draft translations, ii) post-editing. The drafts generated in (i) can then be refined by human translators, ensuring both speed and accuracy. This post-editing procedure reflects human-AI collaboration in MT process (see also Alkodimi et al., 2024). Post-editing can be seen as a modifying procedure for AI translation output aiming to capture the cultural, religious, and register nuances which were missed by AI tools in the initial drafts (cf. Krings, 2001; Groves & Schmidtke, 2009), iii) this collaboration ensures that AI complements human expertise rather than replacing it (see also Jung et al., 2024), and iv) use of hybrid translation models, where AI handles repetitive and straightforward content while human translators focus on nuanced, creative, or culturally sensitive translation. This hybrid technique provides an optimal balance that leverages the strengths of both AI and human expertise (see also Tosun, 2024; Dugonik et al., 2023).

This study thus recommends that future research be conducted on how to overcome difficulties encountering low-resource languages such as Hindi, Bengali, Amharic, Galician, and Basque. Future studies should also focus on H & L dialectical languages including Arabic (cf. Pang et al., 2025). Specifically, Arabic L dialects present real challenges for LLMs, because sometimes they are unintelligible to each other. Cultural as well as religious nuances should be addressed and the same thing can also be done regarding other poorly represented registers. These areas should be the main focus of future studies cross-linguistically so that future training (internet) data involve these missing aspects.

Our study has some limitations, though. First, it involves only articles published in English, which may lead to an incomplete global picture, specifically for regions where AI and translation research is emerging but not published in English. Thus, a broader study could also involve non-English publications. Another limitation that could be highlighted here is that our study considers 3 sources, namely WoS, Scopus, and Lens, though being the most reliable and rigorously indexed databases. A future study could also involve some articles from less indexed sources.

**Competing Interest Declaration**
There are no competing interests to declare.

**Funding Declaration**
This research article did not receive any external or internal funding.

**Data Availability Declaration**
This study does not involve any dataset.

**Ethics, Consent to Participate, and Consent to Publish declarations**: not applicable.

# Annex 1: Sample of materials and analyses

## Research centers (based on number of citations)

Create Map

**Verify selected organizations**

| Selected | Organization | Documents | Citations | Total link strength |
|---|---|---|---|---|
| ☑ | mit | 85 | 3421 | 30028 |
| ☑ | stanford univ | 53 | 2745 | 10625 |
| ☑ | univ calif berkeley | 30 | 2427 | 7979 |
| ☑ | carnegie mellon univ | 61 | 2393 | 27294 |
| ☑ | microsoft corp | 8 | 1609 | 6709 |
| ☑ | chinese acad sci | 140 | 1540 | 94638 |
| ☑ | univ tubingen | 13 | 1530 | 3643 |
| ☑ | harvard univ | 18 | 1409 | 7961 |
| ☑ | univ toronto | 43 | 1377 | 13723 |
| ☑ | univ calif san diego | 22 | 1232 | 9326 |
| ☑ | king saud univ | 23 | 1188 | 10315 |
| ☑ | nanyang technol univ | 18 | 1180 | 10192 |
| ☑ | univ freiburg | 12 | 1068 | 3780 |
| ☑ | univ arizona | 9 | 1039 | 900 |
| ☑ | nyu | 23 | 1026 | 17694 |
| ☑ | univ chinese acad sci | 79 | 989 | 68843 |
| ☑ | broad inst mit & harvard | 18 | 907 | 4927 |
| ☑ | univ calif san francisco | 15 | 874 | 4971 |
| ☑ | harvard med sch | 34 | 866 | 10716 |
| ☑ | yale univ | 15 | 844 | 5236 |

< Back | Next > | Finish | Cancel

## Research centers (based on number of documents)

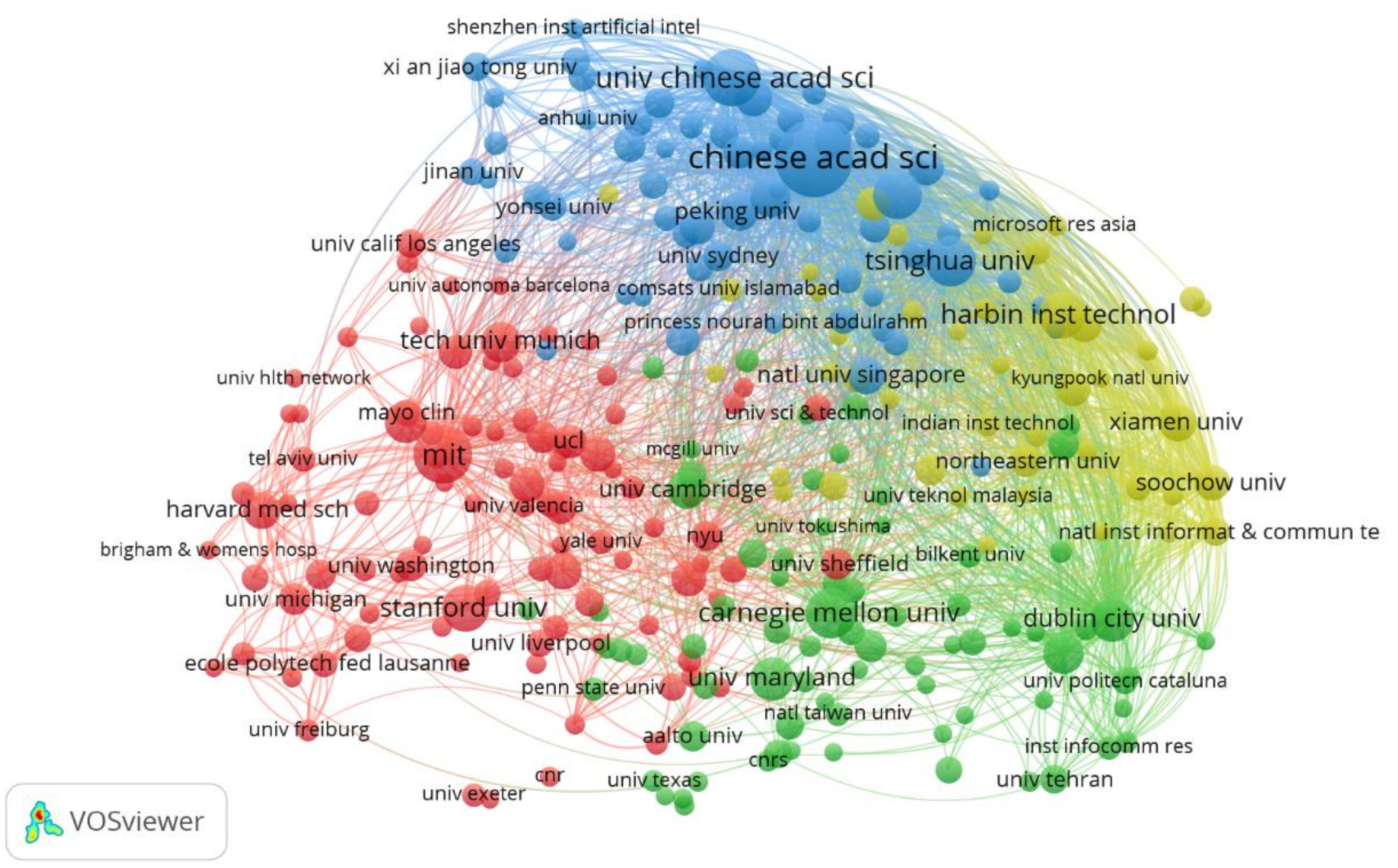

## Journal Materials (sample from Lens)

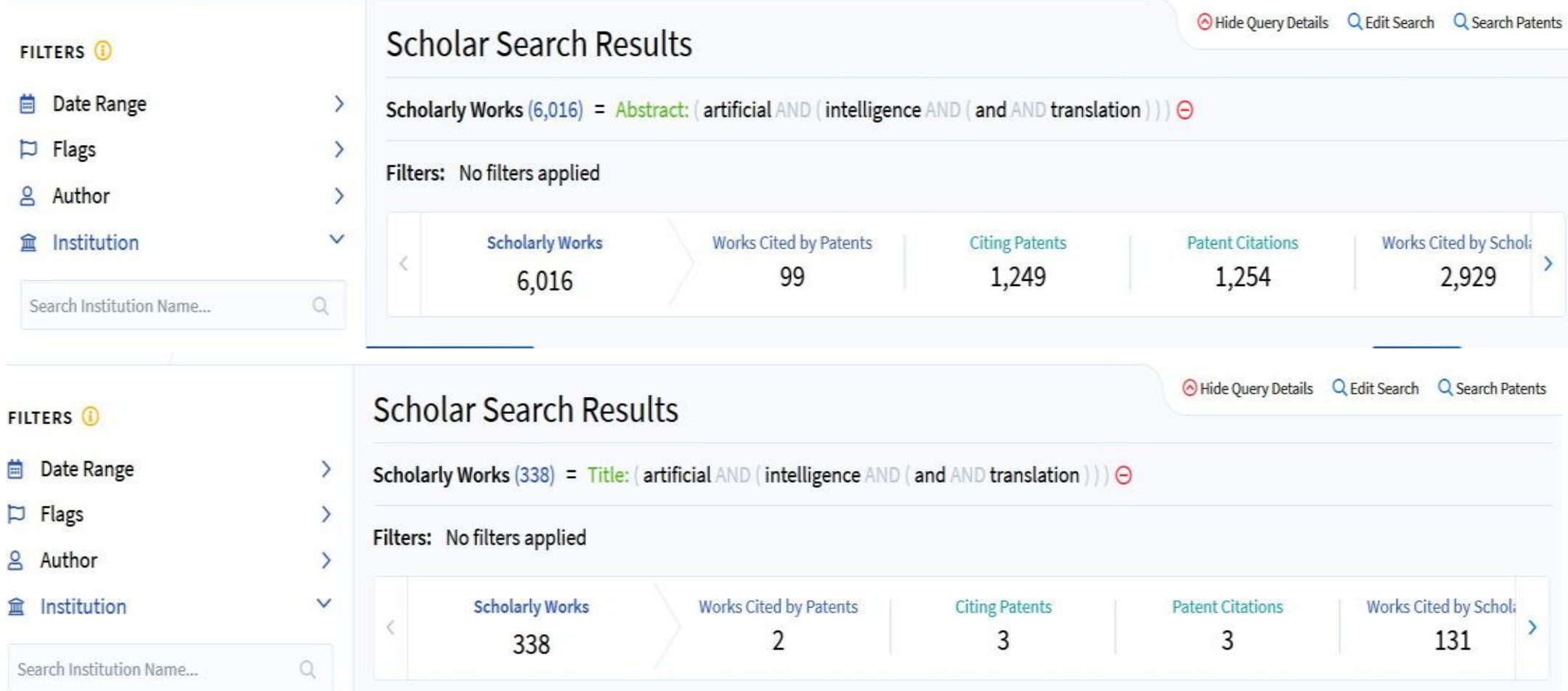

## Cluster view (trending issues)

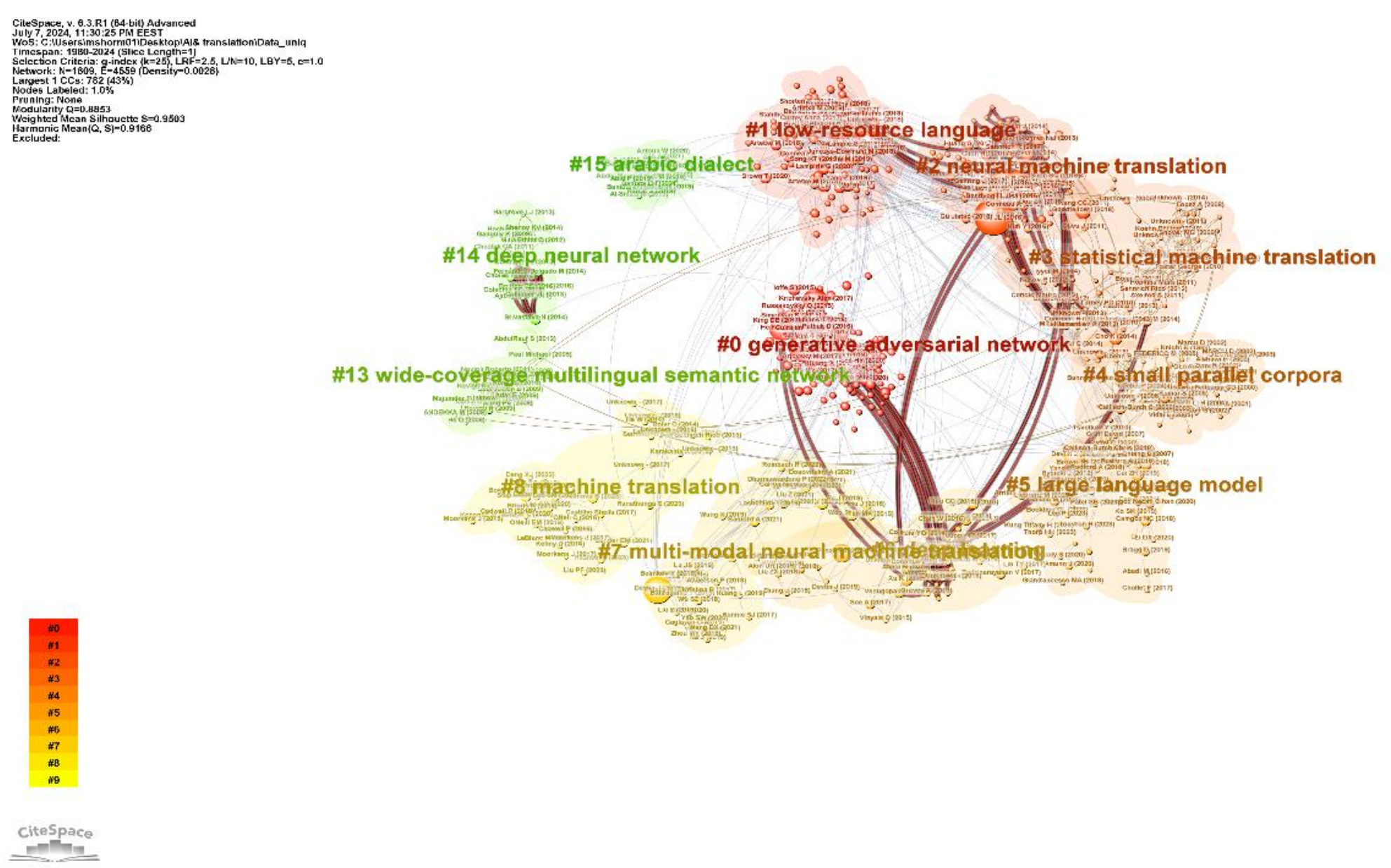

## Total clusters (471)

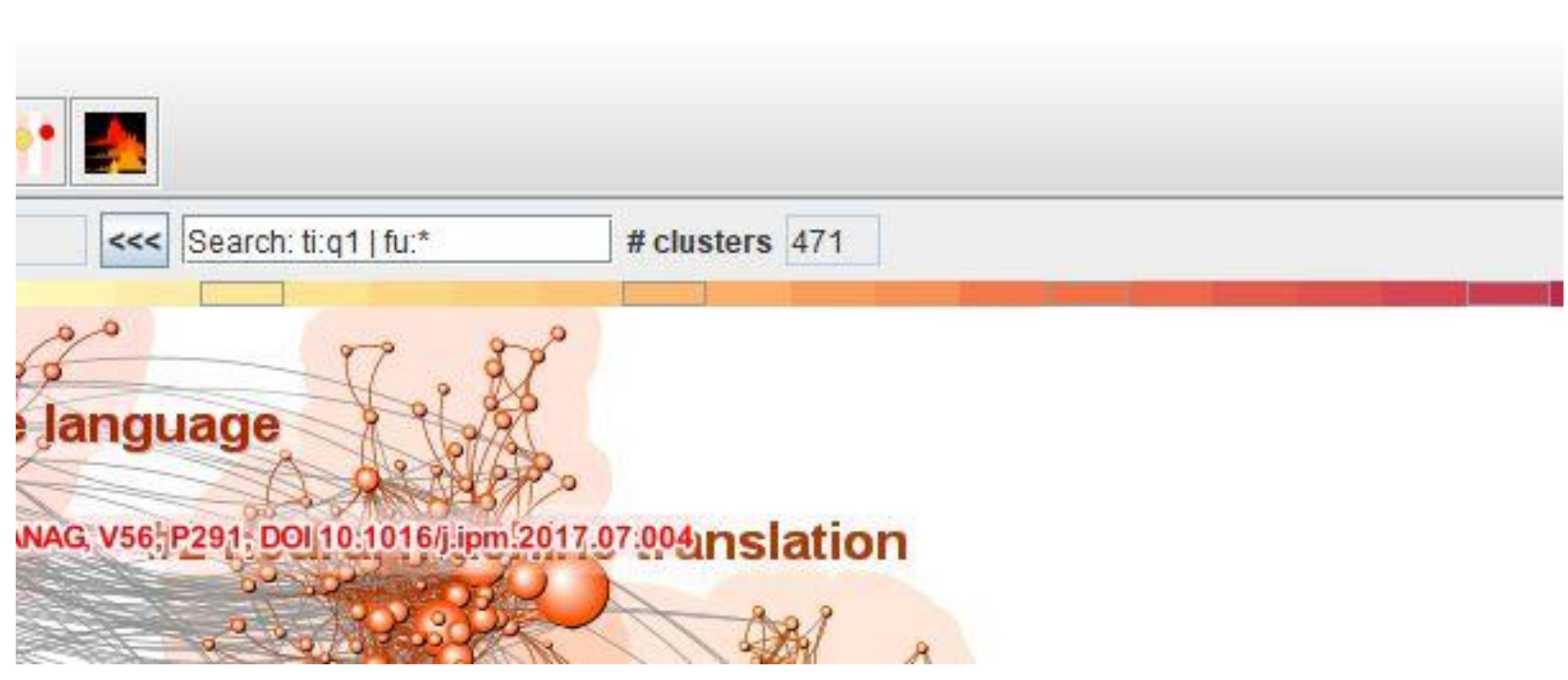